\documentclass[10pt, a4paper, conference, compsocconf]{IEEEtran}
\usepackage[linesnumbered,ruled,vlined]{algorithm2e}
\usepackage{epsfig}
\usepackage{subfigure}
\usepackage{calc}
\usepackage{textcomp}
\usepackage{amssymb}
\usepackage{amstext}
\usepackage{amsmath}
\usepackage{array}

\ifCLASSINFOpdf
\else
\fi
\begin{document}
%
\title{Foreground-Background Segmentation Based on Codebook and Edge\\ Detector}

\author{
\authorblockN{Mika\"el A. Mousse\authorrefmark{1}\authorrefmark{2}, Eug\`ene C. Ezin\authorrefmark{1} and
Cina Motamed\authorrefmark{2}  }
\authorblockA{\authorrefmark{1}Unit\'e de Recherche en Informatique et Sciences Appliqu\'ees\\
Institut de Math\'ematiques et de Sciences Physiques\\
Universit\'e d'Abomey-Calavi, B\'enin\\
BP 613 Porto-Novo\\
Email: \{mikael.mousse, eugene.ezin\}@imsp-uac.org}
\authorblockA{\authorrefmark{2}Laboratoire d'Informatique Signal et Image de la C\^ote d'Opale\\
Universit\'e du Littoral C\^ote d'Opale, France\\
50 rue F. Buisson,
BP 719, 62228 Calais Cedex\\
Email: motamed@lisic.univ-littoral.fr}
}


%


\maketitle

\begin{abstract}
Background modeling techniques are  used for moving object detection in video. Many algorithms
exist in the field of object detection with different purposes. In this paper, we propose an improvement
of moving object detection based on codebook segmentation. We associate the original codebook algorithm 
with an edge detection algorithm. Our goal is to prove the efficiency of using an edge detection
algorithm with a background modeling algorithm. 

Throughout our study, we compared the quality of the moving object
detection when codebook segmentation algorithm is associated with some standard edge detectors. In each case, we use 
frame-based metrics for the evaluation of the detection. The different results are presented and analyzed.

\end{abstract}


%
\IEEEpeerreviewmaketitle

\section{Introduction}
The detection of moving objects in video sequence is the first step in video surveillance system.
The performance of the visual surveillance depends on the quality of the object detection. Many segmentation algorithms
extract moving objects from image/video sequences. The goal of segmentation is to isolate moving objects from
stationary and dynamic background. The variation of local or global light intensity, the object shadow,
the regular or irregular background and foreground  have an impact on the results of object detection. 

The object detection
techniques are subdivided in three categories which are without background modeling, with background modeling and
combined approach. The techniques based on background modeling are recommended in case of dynamic background observed by a static camera.
These techniques generally model the background with
respect to relevant image features. So Foreground pixels can
be determined if the corresponding features from the input image significantly differ from those
of the background model. Three methods are used: background modeling, background estimation, background
substraction. Many research works have already been done \cite{kim,mog,mog1,mog2}. Generally, background 
modeling techniques improve the foreground-background segmentation performance significantly in almost every
challenging environment. They have better performance in both outdoor and indoor environments. The  objects
are integrated in background if they remain static over a specific delay. Sudden variation of light intensity make background model unstable. 

The method
proposed in \cite{kim} has better performance in these situations. Kim et al. \cite{kim} propose a real time 
foreground background segmentation using codebook model. This algorithm works in two
steps which are learning phase and update phase. The learning consists to determine a background model which is 
compared to the input image. The model is updated with new image. 

In this work, we are interested to combine
the algorithm proposed in \cite{kim} with an edge detection algorithm. The main idea is to highlight the boundaries of objects in a scene. 
The using of an edge detector will verify if foreground pixels detected by the codebook algorithm belong to an object or not. 
In this paper we have explored three edge detectors : Sobel operator, Laplace of Gaussian operator and Canny edge detector.

The paper consists of five sections. In Section \ref{reviewcodebook} we
made a review on moving object detection using codebook. In section \ref{proposed} we presented the proposed algorithm for foreground-background
segmentation.
In Section \ref{result} we presented the experimental results and we used some measures to evaluate the performance
of the system.
Finally
in Section \ref{conclude} we ended this work with further directions.

\section{Object Detection Based on Codebook}\label{reviewcodebook}
The basic  codebook background model is proposed in \cite{kim}. This method is widely used for 
moving object detection in case of stationary and dynamic background. 

In this method, each pixel is represented
by a codebook $ \mathcal{C}=\{c_{1}, c_{2},......., c_{L}\}$. The length of codebook is different from one pixel
to another. Each codeword $c_{i}$,  $i=1,........,L$ is represented by a RGB vector $v_{i}$ $ (R_{i},G_{i},B_{i})$ and a 6-tuples
$ aux_{i}=$\{{\em \v{I}}$_{i}$, {\em \^I}$_{i}$, $f_{i}$, $p_{i}$, $\lambda_{i}$, $q_{i}$\} where
\v{I} and \^{I} are the minimum and maximum brightness of all 
pixels assigned to this codeword $c_{i}$, $f_{i}$ is the frequency at which the codeword has occurred, $\lambda_{i}$ is the
maximum negative run length defined as the longest interval during the training period that the codeword has not
recurred, $p_{i}$ and $q_{i}$ are the first and last access times, respectively, that the codeword has occurred.

The codebook is created or updated using two criteria. The first criteria is based on color distorsion $\delta$
whereas the second is based on brightness distorsion. We calcule the color distorsion $\delta$ using equation (\ref{eq}).
 \begin{equation}\label{eq}
    \delta=\sqrt{||p_{t}||^2-C_{p}^2}
 \end{equation}
In this equation, $C_{p}^2$ is the autocorrelation of R, G and B colors of
input pixel $p_{t}$ and the codeword $c_{i}$, normalized by brightness. The autocorrelation value is given by equation (\ref{eq1}).
 \begin{equation}\label{eq1}
    C_{p}^2=\frac{(R_{i}R+G_{i}G+B_{i}B)^2}{R_{i}^2+G_{i}^2+B_{i}^2}
 \end{equation}
According to \cite{kim}, the brightness $I$ has delimited by two bounds. The lower bound is $I_{low}=\alpha${\em \^I}$_{i}$   and the upper limit is
$I_{hi}=min\{\beta${\em \^I}, $ \displaystyle \frac{\text{{\em \v{I}}}}{\alpha}\}$. 
For an input pixel which have R, G and B colors, the formula of the brightness is given by equation (\ref{eq2}).
 \begin{equation}\label{eq2}
    I=\sqrt{R^2+G^2+B^2}
 \end{equation}
For each input pixel, if we find a codeword $c_{i}$ which respect these two criteria (distorsion criteria and brightness criteria) then we update this codeword by setting $v_{i} $ to $ (\frac{f_{i}R_{i}+R}{f_{i}+1},\frac{f_{i}G_{i}+G}{f_{i}+1},\frac{f_{i}B_{i}+B}{f_{i}+1})$
and $  aux_{L} $ to \{$min(I,${\em \v{I}}$_{i}$), $max(I,${\em \^I}$_{i}$), $f_{i}+1$, $max(\lambda_{i},t-q_{i})$, $p_{i}$, $t$\}. If we don't find
a matched codeword, we create a new codeword $c_{K}$. In this case, $v_{K}$ is equal to $(R,G,B)$ and $  aux_{K}$ is equal to \{$I$, $I$, $1$, $t-1$, $t$, $t$\}.

After the training period, if an incoming pixel matches to a codeword in the codebook, then this codeword is
updated. If the pixel doesn't match, his information is put in cache word and this pixel is treated as a foreground
pixel. If a cache word is matched more frequently so this cache word is put into codebook. 

Although the original codebook is a robust background modeling technique, there
are some failure situations. Firstly, for example, in winter, people commonly
use black coats. If foreground-background segmentation is done using the codebook method, it may adopt
black colour as background for many pixels. That is why
a lot of pixels are incorrectly segmented. Secondly, if an object in the scene stops its motion, then it is absorbed in
the background. Kim et al. \cite{kim} indicate tuning parameter  to
overcome this problem, but these modifications reduce the global performance of the algorithm in another situation. Due
to the performance of the proposed method by \cite{kim}, several researchers continue by digging further.
These improvements can be classified into four points.

The first point is the improvement of the algorithm suggested by Kim et al. \cite{kim} by changing algorithm's parameters.
In this category, Ilyas et al. \cite{atif} proposed to use maximum negative run length $\lambda$ and 
frequency $f_{i}$ to decide whether to delete codewords or not. They also proposed to move cache codeword 
into the codebook when access frequency $f_{i}$ is large. In \cite{cheng},  Cheng et al. suggested to convert pixels 
from RGB to YUV space. After this conversion they use the V component to build single gaussian model, making 
the whole codebook. Shah et al \cite{shah} used a statistical parameter estimation method to control adaptation
procedure. Pal et al. \cite{amit} spreaded codewords along boundaries of the neighboring layers.
According to this paper, pixels in dynamic region will have more than one
codeword.

The second point is about the improvement of the codebook algorithm by changing algorithm's model. Some papers
such as \cite{anup,huo} are proposed in this category. Doshi et al. \cite{anup} proposed to use the V component 
in HSV representation of pixels to represent the brightness of these pixels. They suggested an hybrid cone-cylinder
model to build the background model. Donghai et al. \cite{huo} proposed codebook background
modeling algorithm based on principal component analysis (PCA). The model overcomes the mistake of gaussian mixture
model sphere model and codebook cylinder model.

The third point concerns the improvement of codebook algorithm by extension on pixels. Doshi et al. \cite{anup}
proposed to convert pixel from RGB to HSV color space and Wu et al. \cite{wu} suggested to extend codebook in both temporal
and spatial dimensions. Then the proposed algorithm  in \cite{wu} is based on the context information. Fang et al. \cite{fang}
 proposed to convert pixels from RGB to HSL color space, and use L component as brightness value to reduce 
amount of calculation.

The fourth point concerns the improvement of the algorithm proposed in \cite{kim} by combining it with other methods.
In this category, some papers are proposed such as \cite{li,yu,liB}. Li et al. \cite{li} suggested to combine
gaussian mixture model and codebook whereas Wu et al. \cite {yu} proposed to combine local binary pattern
(LBP) with codebook to detect object. LBP texture information is used to establish the first layer of
background. Li et al \cite{liB} proposed to use single gaussian to model codewords. It builds a texture-wise background
model by LBP. This work proposes moving object detection based on the combination of codebook with edge detector. We use the 
gradient information of  the pixel to improve the detection.
\section{Proposed Algorithm}\label{proposed}
Our proposed algorithm consists to combine the codebook with an edge detector algorithm. The goal of this combination is
to improve the moving object detection in video.


After running the codebook algorithm for foreground-background segmentation we proposed to find the convex hull of each contour which have been
detected in the result. We computed an edge detection algorithm and applied it on the original frame which have been converted from
color image to grayscale.
We perform a two-level thresholding.
We thresholded image by using the edge detector response and pixels are displayed only if the gradient is greater than a value $\varphi$.
The value of $\varphi$ is given by formula (\ref{phi}).
\begin{equation}\label{phi}
    \varphi=G(1-\theta)
 \end{equation}
In equation (\ref{phi}), $G$ is the maximum gradient of input image and $\theta$ is a variable which value belongs to [0 1].
The value of $\theta$ depends on the characteristics of the input sequence.
This double thresholding allows us to select only the major edges. After that we also find the convex hull of contours which have been detected in thresholded image.
At this step the potential objects which are on the frame are detected.
Finally a comparison between the pixels detected by codebook and the pixel detected once thresholding is done. 
The role of this comparison is to identify effective foreground pixels. An effective foreground pixel is a pixel which has been
classified to foreground pixels by codebook and has been detected to be an object's pixel by the edge detector.

The detailed algorithm is 
given by Algorithm 1. In this algorithm, we assume that :
\begin{itemize}
 \item for each image of the sequence, the result of the segmentation is given by $r$;
 \item the input pixel $p_{t}$ has R, G and B colors;
 \item $N$ is the number of images that we use for the training;
 \item $L$ is the length of codebook;
 \item the size of the input image $F_{t}$ is $m\times n$;
 \item $\varPsi$, $t_{1}$, $t_{2}$, $r$, $\varpi$ are the grayscale images which have same size with initial image $F_{t}$.
  \item $Threshold$ $(x)$  is a procedure which thresholds the image $x$. The detailed procedure is given in Algorithm 2;
 \item $BGS(F_{t})$  is a procedure which subtracts the current image $F_{t}$ from the background model. It's described in \cite{kim}. For all pixels of the frame $F_{t}$,
 this procedure searches a matched codeword in the codebook. If a pixel doesn't match to any codeword, this pixel is treated as a foreground pixel.
 
\end{itemize}
\begin{figure}[!t]

 \begin{tabular}{m{2cm} m{2cm} m{2cm}}
 \\
 \includegraphics[scale=0.21]{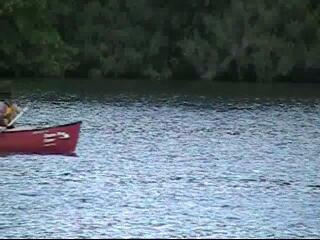}&\includegraphics[scale=0.21]{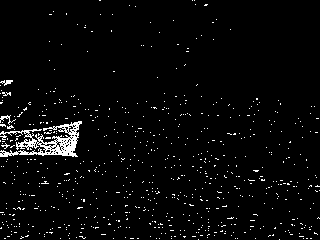}&\includegraphics[scale=0.21]{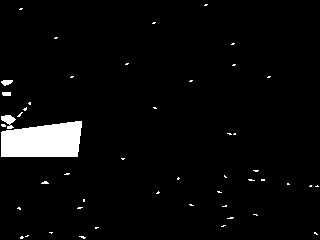}\\
 \multicolumn{1}{c}{\small{(a)}}&\multicolumn{1}{c}{\small{(b)}}&\multicolumn{1}{c}{\small{(c)}}\\
 \includegraphics[scale=0.21]{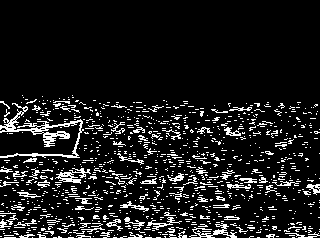}&\includegraphics[scale=0.21]{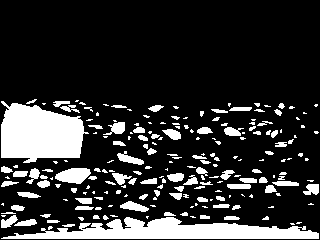}&\includegraphics[scale=0.21]{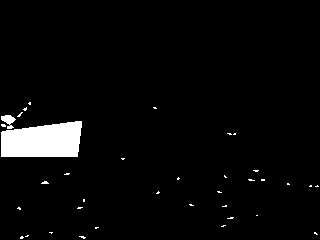}\\
 \multicolumn{1}{c}{\small{(d)}}&\multicolumn{1}{c}{\small{(e)}}&\multicolumn{1}{c}{\small{(f)}}\\
 \end{tabular}
  \caption{(a) image $F_{t}$, (b) image $\varPsi$, (c) image $t_{1}$, (d) image $\varpi$ (using Sobel operator), (e) image $t_{2}$, (e) image $r$.}
 \label{step}
\end{figure}

\begin{algorithm}
\DontPrintSemicolon
\KwIn{video sequence $S$}
\KwOut{moving object}
$l \leftarrow 0, t \leftarrow 1$ \;
\For{each frame $F_{t}$ of input sequence $S$}{
  \For{each pixel $p_{t}$ of frame $F_{t}$}{
    $ p_{t}=(R,G,B)$\;
    $I \leftarrow \sqrt{R^2+G^2+B^2} $\;
    \For{$i=1$ to  $l$}{
    \If{(colordist ($p_{t},v_{i}$)) and (brightness ($I$, \{$\check{I}_{i}$, $\hat{I}_{i}$\}))}{
      Select a matched codeword $c_{i}$\;
      Break\;
    }}
     \eIf{there is no match}{
     $l \leftarrow l+1 $\;
     \begin{small} create codeword $c_{L} $ by setting parameter $v_{L} \leftarrow (R,G,B)$ and $aux_{L} \leftarrow $\{$I$, $I$, $1$, $t-1$, $t$, $t$\} \end{small}\;
    }
     {
      update codeword $c_{i} $ by setting\\
    $v_{i} \leftarrow (\frac{f_{i}R_{i}+R}{f_{i}+1},\frac{f_{i}G_{i}+G}{f_{i}+1},\frac{f_{i}B_{i}+B}{f_{i}+1})$  and 
    \begin{small}
$  aux_{i} \leftarrow$ \{$min(I,$ ${\check{I}}_{i}$), $max(I,$ $\hat{I}_{i}$), $f_{i}+1$, $max(\lambda_{i},t-q_{i})$, $p_{i}$, $t$\}
\end{small}\;
    }
  }
  \For{each codeword $c_{i}$}{
  $\lambda_{i} \leftarrow max\{\lambda_{i}, ((m \times n \times t)-q_{i}+p_{i}-1)\}$\;}
  \If{$ t>N $}{
  $\varPsi \leftarrow BGS(F_{t})$\;
  \begin{small} set the pixels which are in the convex hull of each contour detected in $\varPsi$  to foreground pixel $(t_{1})$. \end{small}\;
  \begin{small} $ \varpi \leftarrow Threshold$ (convert $F_{t}$ from RGB to grayscale)\end{small}\;
  \begin{small} set the pixels which are in the convex hull of each contour detected in  $\varpi$  to foreground pixel $(t_{2})$. \end{small}\;
  $r \leftarrow Intersect$\footnotemark $(t_{1},t_{2})$\;
  }
 $t \leftarrow t+1$ \;
}

\caption{Foreground-background segmentation}
\label{algo:duplicate2}
\end{algorithm}
\newpage
\begin{algorithm}
\DontPrintSemicolon 
\KwIn{grayscale image $G$}
\KwOut{thresholded image $t$}
$t \leftarrow$ detectedge\footnotemark (G)\;
\For{each pixel $p_{i}$ of $t$}{
  \eIf{intensity of $p_{i}\geq$ maximum of $t$'s pixel intensity$\times$(1$-\theta$)} {
     intensity of $p_{i} \leftarrow 255 $\;
  }{
   intensity of $p_{i} \leftarrow 0 $
   }
}
\caption{procedure  $Threshold$  }
\label{algo:change}

\end{algorithm}
The figure \ref{step} illustrates results obtained by the intermediate steps of the proposed algorithm.

\footnotetext[1]{$Intersect$ $(t_{1},t_{2})$ returns an image. The result's pixels are considered to be foreground pixels if the corresponding pixels are considered as foreground pixels both by $t_{1}$ and $t_{2}$.}

\section{Experimental  Results and Performance}\label{result}

In this section,  we present the performance of the
proposed approach by comparing with the codebook algorithm \cite{kim} and mixture of gaussian algorithm \cite{mog}.
The section consists on  two subsections.  The first subsection presents the experimental results whereas
the second presents and analyzes the performance of each algorithm.

\subsection{Experimental Results}
\begin{figure*}[!t]

 \begin{tabular}{m{1cm} m{3.3cm} m{3.3cm} m{3.5cm} m{3.5cm}}
 \\
 Original Frame&\includegraphics[scale=0.30]{in000872.jpg}&\includegraphics[scale=0.30]{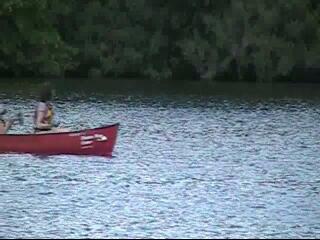}&\includegraphics[scale=0.25]{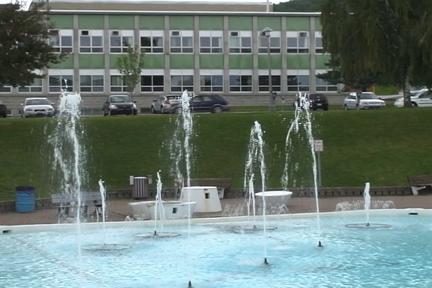}&\includegraphics[scale=0.25]{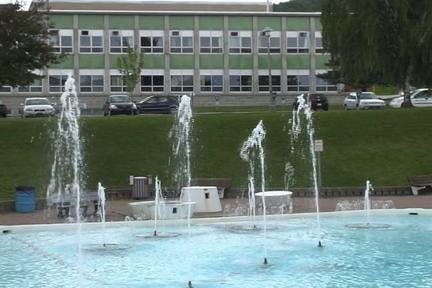}\\
 Ground Truth&\includegraphics[scale=0.40]{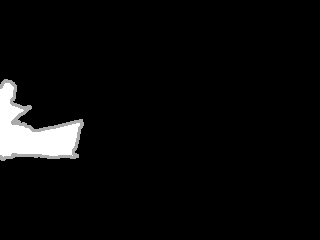}&\includegraphics[scale=0.40]{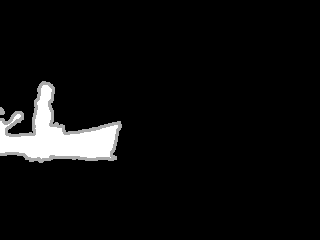}&\includegraphics[scale=0.33]{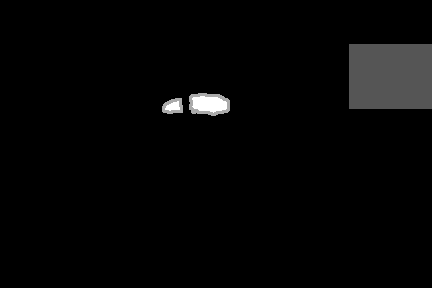}&\includegraphics[scale=0.33]{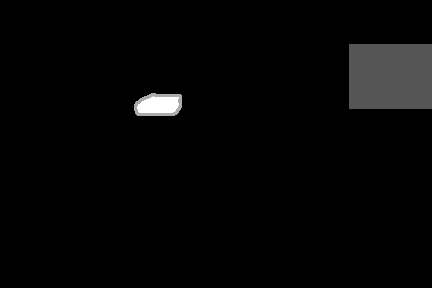}\\
 CB results&\includegraphics[scale=0.30]{cb000872.png}&\includegraphics[scale=0.30]{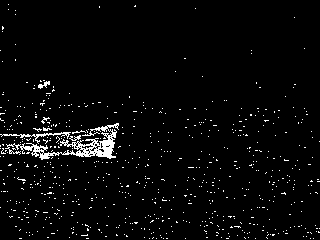}&\includegraphics[scale=0.25]{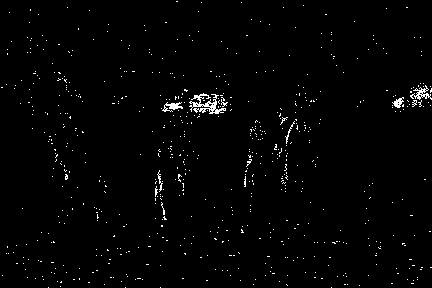}&\includegraphics[scale=0.25]{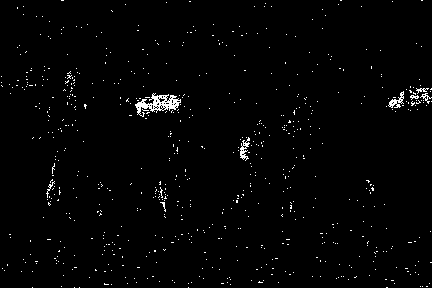}\\
 MoG results&\includegraphics[scale=0.30]{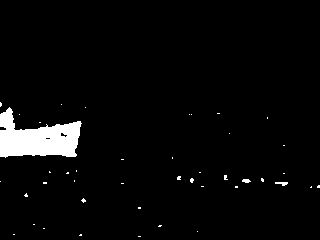}&\includegraphics[scale=0.30]{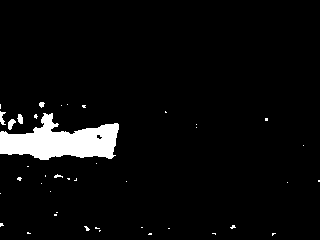}&\includegraphics[scale=0.25]{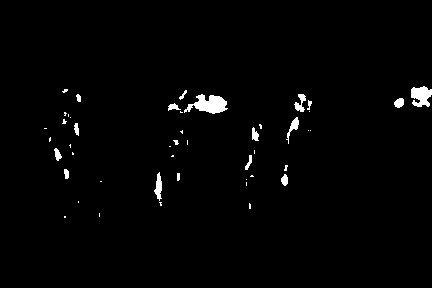}&\includegraphics[scale=0.25]{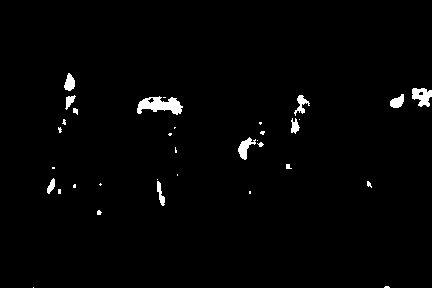}\\
 MCBSb results&\includegraphics[scale=0.30]{sb000872.png}&\includegraphics[scale=0.30]{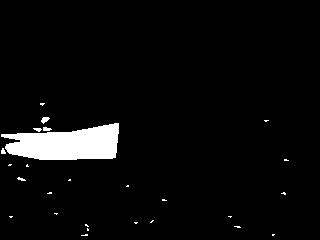}&\includegraphics[scale=0.25]{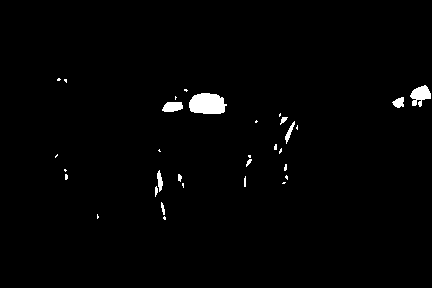}&\includegraphics[scale=0.25]{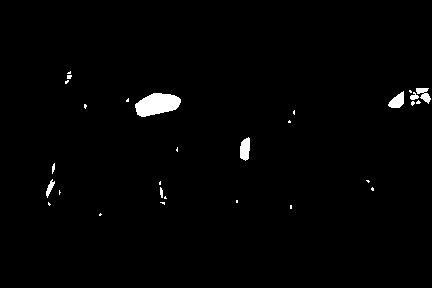}\\
 MCBLp results&\includegraphics[scale=0.30]{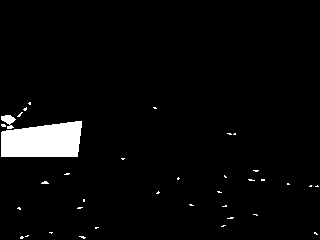}&\includegraphics[scale=0.30]{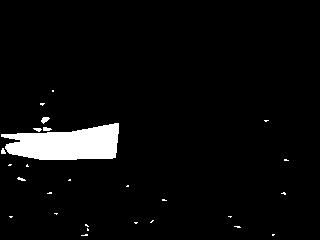}&\includegraphics[scale=0.25]{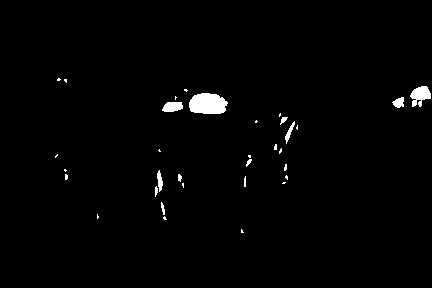}&\includegraphics[scale=0.25]{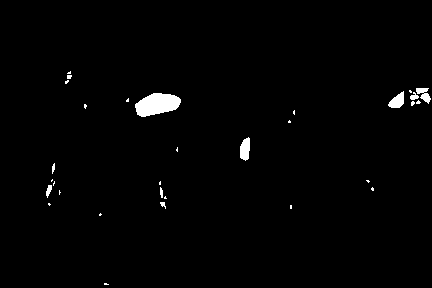}\\
 MCBCa results&\includegraphics[scale=0.30]{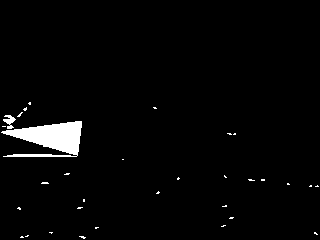}&\includegraphics[scale=0.30]{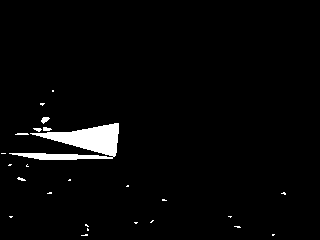}&\includegraphics[scale=0.25]{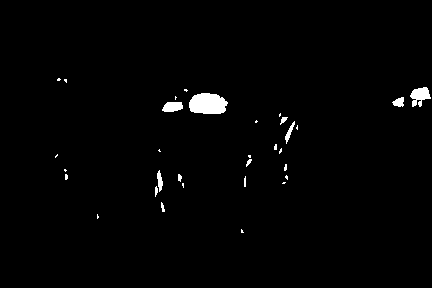}&\includegraphics[scale=0.25]{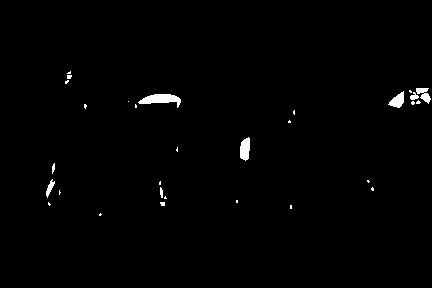}\\
 \end{tabular}
  \caption{Segmentation results}
 \label{fig}
\end{figure*}

For the validation of our algorithm, we have selected two benchmarking datasets from 
\cite{changelien} covered under the work done by \cite{changepaper}. They are ``Canoe'' and ``fountain01'' datasets.
The experiment environment is IntelCore7@2.13Ghz processor with 4GB memory and the programming language is C++.
The parameters
settings for mixture of gaussian were $\alpha$ = 0.01, $\rho$ = 0.001, K = 5, T = 0.8 and $\lambda$ = 2.5$\sigma$.
These parameters were suggested in \cite{mog}. According to \cite{kim}, for codebook, parameter $\alpha$ is between 0.4 and
0.7, and parameter $\beta$ belongs [1.1  1.5]. In this work, we take $\alpha= 0.4$ and $\beta = 1.25$. The parameter
$\theta$ of our proposed method depends on the dataset. For ``Canoe'' dataset we use $\theta=0.85$ whereas for 
``fountain01'' dataset we use $\theta=0.80$. The results of segmentation are given in Figure \ref{fig}.
We assume that :
\begin{itemize}
 \item CB means codebook;
 \item MoG means mixture of gaussian;
 \item MCBSb means combination of codebook and Sobel;
 \item MCBLp means combination of codebook and Laplacian of Gaussian operator;
 \item MCBCa means combination of codebook and Canny edge detector.
\end{itemize}

\subsection{Performance Evaluation and Discussion}
\footnotetext[2]{detectedge refers to the edge detector (Sobel operator, Laplace of Gaussian operator and Canny edge detector).}

In each case, we use an evaluation based on ground truth to show the performance of the segmentation algorithm.
The ground truth has been obtained by labelling objects of interest in the original frame.

The ground truth
based metrics are : true negative (TN), true positive (TP), false negative (FN) and false positive (FP). A pixel
is a true negative pixel when both ground truth and system result agree on the absence of object. A pixel
is a true positive pixel when ground truth and system agree on the presence of objects. A pixel is a false
negative (FN) when system result agree of absence of object whereas ground truth agree of the presence of  object.
A pixel is a false positive (FP) when the system result agrees with the presence of object whereas ground truth agree
with the absence of object.
With these metrics, we compute other parameters which are:
\begin{itemize}
 \item {\bf False positive rate} (FPR) using formula (\ref{fpr});
 \begin{equation}\label{fpr}
    FPR=1-\frac{TN}{TN+FP}
\end{equation}
\item {\bf True positive rate} (TPR) using formula (\ref{tpr});
 \begin{equation}\label{tpr}
    TPR=\frac{TP}{TP+FN}
\end{equation}
\item {\bf Precision} (PR) using formula (\ref{pr});
 \begin{equation}\label{pr}
    PR=\frac{TP}{TP+FP}
\end{equation}
\item {\bf F-measure} (FM) using formula (\ref{fm});
 \begin{equation}\label{fm}
    FM=\frac{2\times PR\times TPR}{PR+TPR}
\end{equation}
\end{itemize}
We also compare the segmentation methods by using {\bf percentage of correct classification} (PCC) 
and {\bf Jaccard coefficient}
(JC). PCC is calculate with formula (\ref{pcc}) and JC is calculate with formula (\ref{jc}).
\begin{equation}\label{pcc}
    PCC=\frac{TP+TN}{TP+FN+FP+TN}
\end{equation}
\begin{equation}\label{jc}
    JC=\frac{TP}{TP+FP+FN}
\end{equation}
We present the results in Table \ref{rec} and Table \ref{rec1}.
\begin{table}[!t]
\renewcommand{\arraystretch}{1.5}
 \caption{Comparison of different metrics according to experiments with dataset ``canoe''}
 \label{rec}
 \begin{tabular}{|c|c|c|c|c|c|}
 \hline
 \normalsize{Metrics}&\normalsize{CB}&\normalsize{MoG}&\normalsize{MCBSb}&\normalsize{MCBLp}&\normalsize{MCBCa}\\
 \hline
 \normalsize{FPR}  &1.62&0.36&0.29&0.31&{\bf 0.24}\\
 \hline
 \normalsize{PR}&41.01&{\bf 89.82}&86.29&86.01&81.89\\
 \hline
 \normalsize{FM}&35.08&{\bf 88.17}&63.08&65.09&43.39\\
 \hline
 \normalsize{PCC} &95.98&{\bf 99.18}&97.94&98.01&97.27\\
 \hline
 \normalsize{JC} &21.27&{\bf 78.85}&46.07&48.24&27.71\\
 \hline
 \end{tabular}
\end{table}
\begin{table}[!t]
\renewcommand{\arraystretch}{1.5}
 \caption{Comparison of different metrics according to experiments with dataset ``fountain01''}
 \label{rec1}
 \begin{tabular}{|c|c|c|c|c|c|}
 \hline
 \normalsize{Metrics}&\normalsize{CB}&\normalsize{MoG}&\normalsize{MCBSb}&\normalsize{MCBLp}&\normalsize{MCBCa}\\
 \hline
 \normalsize{FPR}  &1.09&1.59&{\bf 0.43}&0.45&{\bf 0.43}\\
 \hline
 \normalsize{PR}&2.24&4.01&6.82&{\bf 7.31}&6.79\\
 \hline
 \normalsize{FM}&4.17&7.63&11.58&{\bf 12.50}&11.53\\
 \hline
 \normalsize{PCC} &98.86&98.40&{\bf 99.51}&99.49&{\bf 99.51}\\
 \hline
 \normalsize{JC} &2.13&3.97&6.14&{\bf 6.67}&6.11\\
 \hline
 \end{tabular}
\end{table}
In these
Tables, we assume that, value in bold are the optimal value of the row.
We  analyzied the results in Table \ref{rec} and Table \ref{rec1} throught two steps. At first, we make a comparison between codebook,
mixture of gaussian and our method based on the combination of codebook and edge detector. At the second stage, we make a comparative study
of the performance of the system obtained after combination of codebook with the three edge detectors.

All experiments confirm that when codebook is combined with an edge detector, we get better result than original codebook.
Experiments with ``canoe'' dataset proove that mixture of gaussian has good result than codebook. According to our results, the choice between mixture of gaussian and
our method based on the combination of codebook and edge detector depends upon the application and the dataset's characteristics. 
Experimentals Results with dataset ``fountain01'' show that our method is better than the mixture of gaussian approach. However, according to results of experiments
with dataset ``Canoe'' we have two cases :
\begin{enumerate}
 \item if we want to minimize the
false alarms then FPR should be minimized. In this case, experiments show that method based on the combination of codebook and edge detector
has the best result;
\item if we don't want to miss any foreground pixel we need to maximize TPR and FM. In this case, Experiments with 
allow us to use mixture of gaussian for segmentation.
\item[]
\end{enumerate}
The experimental results also proove that the choice of edge detection algorithm depends upon the application.
For example, for an application in which the real-time parameter is not important, the use of Canny operator is recommended if 
we want to minimize the false alarms. But if we need to improve TPR then we use a Laplacian of Gaussian operator. If we want to make
a real-time application, we need to use Sobel operator, because the complexity of Sobel operator is less than Laplacian of Gaussian operator and 
Canny operator. The using of Sobel operator increases the codebook algorithm processing time by 19.55\% (23.33\% for laplacian
of Gaussian and 28.15\% for canny edge detector).
\section{Conclusion}\label{conclude}
In this paper, we present a novel algorithm to segment moving objects with an approach combining the codebook and edge detector. Firstly, we segment sequence
using codebook algorithm. This segmentation help us to know background pixels and foreground pixels. After that, by using edge detector, we show
the object boundaries in each sequence. Then we set all foreground pixels which are not object's pixel to background pixel. The results can be summarized as follow :
\begin{itemize}
 \item our method outperforms the codebook algorithm \cite{kim} in accuracy;
 \item in \cite{kim}, authors claimed that codebook algorithm works better than the mixture of gaussian algorithm. This is not always true;
 \item the choice between our algorithm and the mixture of gaussian algorithm \cite{mog} depends on the input dataset's characteristics
and the final application;
 \item the choice of edge detection algorithm which combines with codedook algorithm depends also on  the characteristics
of the sequence and the final application .
\end{itemize}
In the future,
we will propose an extended version by adding a region based information in order to improve the compactness
of the foreground object.

\section*{Acknowledgment}
This work is partially financially supported by the Association AS2V and Fondation Jacques De Rette, France. We are also grateful 
to Professor Kokou Yetongnon for his fruitful comments.

\end{document}